\documentclass{article}

\usepackage{arxiv}
\usepackage[utf8]{inputenc} 
\usepackage[T1]{fontenc}    
\usepackage{hyperref}       
\usepackage{url}            
\usepackage{booktabs}       
\usepackage{amsfonts}       
\usepackage{nicefrac}       
\usepackage{microtype}      
\usepackage{lipsum}
\usepackage{graphicx} 

\newcommand{\etal}{\textit{et al.}}
\usepackage{adjustbox}
\usepackage{tikz,pgfplots}

\usepackage{caption}
\usepackage{subcaption}
\usepackage{float}

\usepackage{xspace}

\newenvironment{customlegend}[1][]{%
    \begingroup
    \csname pgfplots@init@cleared@structures\endcsname
    \pgfplotsset{#1}
}{
    \csname pgfplots@createlegend\endcsname
    \endgroup
}

\def\addlegendimage{\csname pgfplots@addlegendimage\endcsname}

\pgfplotsset{
  /pgfplots/xlabel near ticks/.style={
     /pgfplots/every axis x label/.style={
        at={(ticklabel cs:0.5)},anchor=near ticklabel
     }
  },
  /pgfplots/ylabel near ticks/.style={
     /pgfplots/every axis y label/.style={
        at={(ticklabel cs:0.5)},rotate=90,anchor=near ticklabel}
     }
  }

\newcommand{\F}{\textit{LimeOut}\xspace}

\begin{document}
\title{Making ML models fairer through explanations:\\ the case of LimeOut
\thanks{This research was partially supported by TAILOR, a project funded by EU Horizon 2020 research and innovation programme under GA No 952215, and the Inria Project Lab ``Hybrid Approaches for Interpretable AI'' (HyAIAI)}}

\author{
  Guilherme Alves \\
  Universit\'e de Lorraine,\\
 CNRS, Inria, LORIA, F-54000 Nancy, France\\
 \texttt{guilherme.alves-da-silva@loria.fr} \\
 
 \And
 
  Vaishnavi Bhargava \\
 Universit\'e de Lorraine,\\
 CNRS, Inria, LORIA, F-54000 Nancy, France\\
  \texttt{vaishnavi.bhargava2605@gmail.com} \\
   \And
 Miguel Couceiro \\
  Universit\'e de Lorraine,\\
  CNRS, Inria, LORIA, F-54000 Nancy, France\\
  \texttt{miguel.couceiro@loria.fr} \\
   \And
  Amedeo Napoli \\
  Universit\'e de Lorraine,\\
  CNRS, Inria, LORIA, F-54000 Nancy, France\\
  \texttt{amedeo.napoli@loria.fr} \\
  
}

\maketitle
\begin{abstract}
Algorithmic decisions are now being used on a daily basis, and based on Machine Learning (ML) processes that may be complex and biased. This raises several concerns given the critical impact that biased decisions may have on individuals or on society as a whole. Not only unfair outcomes affect human rights, they also undermine public trust in ML and AI.
In this paper we address fairness issues of ML models based on decision outcomes, and we show how the simple idea of ``feature dropout'' followed by an ``ensemble approach'' can improve model fairness. 
To illustrate, we will revisit the case of ``LimeOut'' that was proposed to tackle ``process fairness'', which measures a model's reliance on sensitive or discriminatory features. Given a classifier, a dataset and a set of sensitive features,  LimeOut first assesses whether the classifier is fair by checking its reliance on sensitive features using ``Lime explanations''. If deemed unfair, LimeOut then applies feature dropout to obtain a pool of classifiers. These are then combined into an ensemble classifier that was empirically shown to be less dependent on sensitive features without compromising the classifier's accuracy.
We  present different experiments on multiple datasets and several state of the art classifiers, which show that LimeOut's classifiers improve (or at least maintain) not only process fairness but also other fairness metrics such as individual and group fairness, equal opportunity, and demographic parity.

 
\end{abstract}

\keywords{Fairness metrics \and Feature importance \and Feature-dropout \and  Ensemble classifier \and LIME explanations }

\section{Introduction}
Algorithmic decisions are now being used on a daily basis and obtained by Machine Learning (ML) processes that may be rather complex and opaque. This raises several concerns given the critical impact that such decisions may have on individuals or on society as a whole. Well known examples include the classifiers which are used to predict the credit card defaulters, including multiple other datasets which may impact the government decisions. These prevalent classifiers are generally known to be biased to certain minority or vulnerable groups of society, which should rather be protected. 
Most of the notions of fairness thus focus on the outcomes of the decision process~\cite{speicher2018unified,zafar2017fairness}.
They are inspired by several  anti-discrimination efforts that aim to ensure that unprivileged  groups (e.g. racial minorities) should be treated fairly. Such issues can be addressed by looking into fairness individually~\cite{speicher2018unified} or as a group~\cite{speicher2018unified,zafar2017fairness}. 
Actually, earlier studies \cite{zemel2013learning,zafar2015fairness} consider individual and group fairness as conflicting measures, and some studies tried to find an optimal trade-off between them. In~\cite{binns2020apparent} the author argues that, although apparently conflicting, they correspond to the same underlying moral concept, thus providing a broader perspective and advocating an individual treatment and assessment based on a case-by-case analysis.
\\
The authors of ~\cite{grgic2016case,grgic2018beyond} provide yet another noteworthy perspective of fairness, namely, \emph{process fairness}. Rather than focusing on the outcome, it deals with the process leading to the outcome.
In~\cite{bhargava} we delivered a potential solution to deal with process fairness in ML classifiers. The key idea was to use an explanatory model, namely, LIME~\cite{lime} to assess whether a given classifier was fair by measuring its reliance on salient or sensitive features.  This component was then integrated in a human-centered workflow called \F,  that receives as input a triple $(M,D,F)$ of a classifier $M$, a dataset $D$ and a set $F$ of sensitive features, and outputs a classifier $M_{final}$ less dependent on sensitive features without compromising accuracy. To  achieve  both  goals, LimeOut  relies  on feature dropout to produce a pool of classifiers that are then combined through an ensemble approach. Feature dropout receives a classifier and a feature $a$ as input, and produces a classifier that does not take $a$ into account. 
This preliminary study~\cite{bhargava} showed the feasibility and the flexibility of the simple idea of feature dropout followed by an ensemble approach to improve process fairness. 
However, the  empirical  study of~\cite{bhargava} was performed only on  two  families  of  classifiers  (logistic  regression  and  random  forests)  and  carried  out  on two  real-life  datasets  (Adult  and German Credit Score). Also, 
it did not take into account other commonly used fairness measures. 
Moreover, in a recent study \cite{dimanov}, Dimanov \etal\,  question the trustfulness of certain explanation methods when assessing model fairness. In fact, they  present a procedure for modifying a pre-trained model in order to manipulate the outputs of explanation methods that are  based on feature importance (FI).  They also observed minor changes in accuracy and  that, even though the pre-trained model was deemed fair by some FI based explanation methods, it may conceal unfairness with respect to other fairness metrics. 
\\
This motivated us to revisit \F's framework to perform a thorough analysis
that follows the tracks of \cite{dimanov} and extends the empirical study of~\cite{bhargava} in several ways: (i) we experiment on many other datasets (e.g., HDMA dataset, Taiwanese Credit Card dataset, LSAC) , (ii) we make use of a larger family of ML classifiers (that include AdaBoost, Bagging, Random Forest (RF), and Logistic Regression (LR)), and (iii) we evaluate \F's output classifiers with respect to a wide variety fairness metrics, namely, disparate impact (DI), disparate mistreatment or equal opportunity (EO), demographic parity (DP), equal accuracy (EA), and predictive equality (PQ).
As it will become clear from the empirical results, the robustness of \F's to different fairness view points is once again confirmed without compromising accuracy. 
\\
The paper is organised as follows. After recalling Lime explanations  and various fairness measures in Subsections~\ref{sssec:LIME} and~ \ref{sssec: Model Fairness}, respectively, we briefly describe  \F's workflow in Subsection~\ref{sssec: LimeOut}. We then present in Section~\ref{sec:experiments} an extended empirical study following the tracks of~\cite{bhargava} and the recent study~\cite{dimanov}. First we quickly describe the datasets used (Subsection~\ref{ssec:ds}) and the classifiers employed (Subsection~\ref{ssec:empericalset}). We then present  the empirical results and the various assessments with respect to the different fairness metrics considered in    Subsection~\ref{sssec: Model Fairness}. We conclude the paper in Section~\ref{sec:conclusion} with some final remarks on ongoing work and perspectives of future research.

\section{Related Work}
\label{sec:related-work}
In this section, we briefly recall LIME (Subsection~\ref{sssec:LIME}),  recall the different metrics used to measure model fairness (Subsection~\ref{sssec: Model Fairness}) and revisit \F's framework (Subsection~\ref{sssec: LimeOut}).

\subsection{LIME - Explanatory Method}\label{sssec:LIME}
Recall that LIME explanations~\cite{lime} (Local Interpretable Model Agostic Explanations) take the form of surrogate linear models, that locally mimic the behavior of a ML model.
Essentially, it tries to find the best possible linear model (i.e. explanation model) which fits the prediction of ML model of a given instance and it's neighbouring points (see below). 

\begin{figure*}[hbt]
\centering
    \includegraphics[width=.7\textwidth]{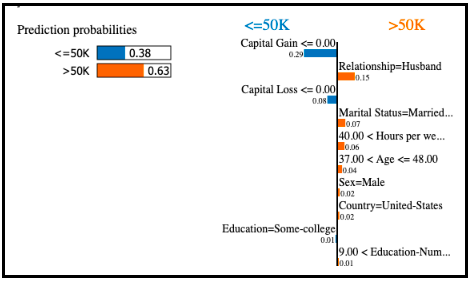}
    \caption{LIME explanation in case of Adult dataset}
    \label{fig:mesh2}
\end{figure*}

Let $f: \mathbb{R}^d \rightarrow \mathbb{R}$ be the function learned by a  classification or regression model over training samples. LIME's workflow can be described as follows.
Given an instance $x$ and its ML prediction $f(x)$, LIME  generates neighbourhood points by perturbing $x$ and gets their corresponding predictions. These neighbouring points $z$ are assigned weights based on their proximity to $x$, using the following equation: 
\begin{equation}
\pi_x(z)=e^{(\frac{-D(x,z)^2}{\sigma^2})},
\end{equation}
where $D(x, z)$ is the Euclidean distance between $x$ and $z$, and $\sigma$ is the hyper parameter (kernel-width).
LIME then learns the weighted linear model $g$ over the original and neighbourhood points, and their respective predictions, by solving the following optimization problem:
\begin{equation}
g = \mathit{argmin}_{g \in \mathcal{G}}{\;\mathcal{L}(f, g, \pi_{x}(z))} + \Omega(g),
\end{equation}
where $L(f,g,\pi_{x}(z))$ is a measure of how unfaithful $g$ is in approximating $f$ in the locality defined by $\pi_{x}(z)$. $\Omega(g)$ measures the complexity of $g$ (regularization term). In order to ensure both interpretability and local faithfulness, LIME  minimizes  $L(f,g,\pi_{x}(z))$ while enforcing $\Omega(g)$ to be small in order to be interpretable by humans. 
The obtained explanation model $g$ is of the form 
$$g(x) = \hat{\alpha}_0 + \sum_{1 \le i \le d'}{\hat{\alpha}_i x[i]},$$ where $\hat{\alpha}_i$ represents the contribution or importance of feature  $x[i]$. Figure \ref{fig:mesh2} presents the  explanation of LIME for the classification of an instance from the Adult dataset. For instance, the value ``Capital Gain''$\leq 0.0$ contributes 0.29 to the class $\leq 50K$, whereas the value ``Relationship''$= Husband$ contributes 0.15 to the class  $> 50K$.

\subsection{Model Fairness}\label{sssec: Model Fairness}

Several metrics have been proposed in the literature in order to assess ML model's fairness. Here we recall some of the most used ones. 

\begin{itemize}
    \item \textbf{Individual Fairness}\footnote{It is also referred to as \textit{disparate treatment} or \textit{predictive parity}}~\cite{chouldechova2017fair} imposes that the instances/individuals belonging to different sensitive groups, but with similar non-sensitive attributes must receive equal decision outcomes.
    
    
    \item \textbf{Disparate Impact\footnote{It is also referred to as \textit{group fairness}}} (DI)
    ~\cite{dwork2012fairness} is rooted in the desire for different sensitive demographic groups to experience similar rates of positive decision outcomes ($\hat{y} = pos$). Given the ML model, $\hat{y}$ represents the predicted class. It compares two groups of the population based on a sensitive feature: the privileged ($priv$) and the unprivileged ($unp$) groups. For instance, if we consider race as sensitive feature, white people can be assigned as privileged and non-white people as unprivileged group.
    \begin{equation}
        DI=\frac{P(\hat{y}=pos|D=unp)}{P(\hat{y}=pos|D=priv)}
    \end{equation}
    
    \item \textbf{Equal Opportunity}\footnote{It is also referred to as \textit{disparate mistreatment}} \textbf{(EO)}~\cite{zafar2017fairness} proposes different sensitive groups to achieve similar rates of error in decision outcomes. It is computed as the difference in recall scores  ($\frac{TP_i}{TP_i+FN_i}$), where $TP_i$ is true positive and $FN_i$ is false negative for a particular group $i$) between the unprivileged and privileged groups.
    
    \begin{equation}
        EO=\frac{TP_{unp}}{TP_{unp}+FN_{unp}}-\frac{TP_{priv}}{TP_{priv}+FN_{priv}}
    \end{equation}
    
    \item \textbf{Process Fairness\footnote{It is also referred to as  \textit{procedural fairness}}}~\cite{grgic2016case,grgic2018beyond} deals with the process leading to the prediction and keeps track of input features used by the decision model. In other words, the process fairness deals at the algorithmic level and ensures that the algorithm does not use any sensitive features while making a prediction.
    
    \item \textbf{Demographic Parity (DP)}~\cite{NIPS2016_6374} the difference in the predicted positive rates between the unprivileged and privileged groups.
    \begin{equation}
        DP=P(\hat{y}=pos|D=unp)-P(\hat{y}=pos|D=priv)
    \end{equation}
    
    \item \textbf{Equal Accuracy (EA)}~\cite{NIPS2016_6374} the difference in accuracy score ($\frac{TP_i+TN_i}{P_i+N_i}$), where $TN_i$ is true negative of a particular group $i$) between unprivileged and privileged groups.

    \begin{equation}
        EA=\frac{TP_{unp}+TN_{unp}}{P_{unp}+N_{unp}}-\frac{TP_{priv}+TN_{priv}}{P_{priv}+N_{priv}}
    \end{equation}
\end{itemize}

\begin{itemize}
    \item \textbf{Predictive Equality (PE)} which is defined as the difference in false positive rates ($\frac{FP_i}{FP_i+TP_i}$), where $FP_i$ is false positive for a particular group $i$) between unprivileged and privileged groups. Formally,
    \begin{equation}
        PE=\frac{FP_{unp}}{FP_{unp}+TP_{unp}}-\frac{FP_{priv}}{FP_{priv}+TP_{priv}}.
    \end{equation}      
\end{itemize}  

In this paper we follow the same empirical setting  of \cite{dimanov} and \cite{bhargava} and, hence, will focus mainly on disparate impact, equal opportunity, process fairness, demographic parity and equal accuracy. 

\subsection{\F's framework}\label{sssec: LimeOut}
In this subsection, we briefly describe \F's framework, which essentially consists of two main components: LIME\textsubscript{Global} and ENSEMBLE\textsubscript{Out}. Given an input $(M,D,F)$, where $M$ is a classifier, $D$ is a dataset, and  $F$ is a list of sensitive features,  \F first employs a ``global variant'' of LIME (LIME\textsubscript{Global}) to assess the contribution (importance) of each feature to the classifier’s outcomes. For that, LIME\textsubscript{Global}  uses submodular pick to select instances with diverse and non-redundant explanations \cite{lime}, and which are then aggregated to provide a global explanations (see \cite{bhargava}). The final output of LIME\textsubscript{Global} is thus a list of the $k$ most important features\footnote{In \cite{bhargava} $k$ was set to 10.}.  
\\
If the $k$ most important feature contain at least two sensitive features in $F$, then the model is deemed unfair (or biased), and the second component ENSEMBLE\textsubscript{Out} is deployed. Essentially, ENSEMBLE\textsubscript{Out} applies feature dropout on the sensitive features that are among the $k$ most important features, each of which giving rise to a classifier obtained from $M$ by removing that feature. thus resulting in a pool of classifiers. ENSEMBLE\textsubscript{Out} then constructs an ensemble classifier $M\textsubscript{final}$ through a linear combination of the pool's classifiers.
\\
More precisely, if LIME\textsubscript{Global} outputs $a_1, a_2,\ldots, a_k$ as the $k$ most important features, in which $a_{j_1}, a_{j_2},\ldots, a_{j_i}$ are sensitive, then \F trains $i+1$ classifiers: $M_t$ after removing $a_{j_t}$ from the dataset, for $t=1, \ldots, i$, and $M_{i+1}$ after removing all sensitive features $a_{j_1}, a_{j_2},\ldots, a_{j_i}$.
The ensemble classifier $M_{final}$ is then  defined as the ``average'' of these $i+1$ classifiers, i.e., by the rule: for an instance $x$ and a class $C$,
\[
{P}_{M_{final}}(x\in C) =  \frac{\sum_{t=1}^{i+1} {P}_{M_t}(x\in C)}{i+1}.
\]
The empirical studies carried out in \cite{bhargava} showed that this ensemble classifier obtained by \F is fairer with respect to process fairness than the input model $M$, without compromising (or even improving) $M$'s accuracy. 


\section{Empirical study}
\label{sec:experiments}

In this section,  we first describe in Subsection \ref{ssec:ds} the datasets  that we used in our experiments, and we briefly present in Subsection \ref{ssec:empericalset} the empirical setup. We then discuss our results from different points of view. In Subsection \ref{sec:acc-assessment} we report on the improved accuracy of \F's classifiers using different models and on the various datasets considered. 
We will then assess the fairness of \F's classifiers in Subsection \ref{sec:acc-assessment}:  first on process fairness and then on the remaining metrics of Subsection~\ref{sssec: Model Fairness}.

\subsection{Datasets}\label{ssec:ds}

Experiments were conducted using five datasets. All datasets share common characteristics that allow us to run our experiments: a binary target feature and the presence of sensitive features. Table \ref{tb:datasets} summarizes basic information about these datasets. The details concerning each dataset are presented as follows.

\begin{table}
\caption{Datasets employed in the experiments.} 
\begin{center}
\begin{tabular}{r|lll}
\hline
Dataset           & \# features & \# sensitive & \# instances \\
\hline
Adult             &  \quad 14          & \quad 3            & \quad 32561        \\
German            & \quad 20          & \quad 3            & \quad 1000         \\
HMDA              & \quad 28          & \quad 3            & \quad 92793        \\
Default           & \quad 23          & \quad 3            & \quad 30000        \\
LSAC              & \quad 11          & \quad2            & \quad 26551        \\
\hline
\end{tabular}   
\end{center}
\label{tb:datasets}
\end{table}

\textbf{Adult.} This dataset is available on UCI repository\footnote{http://archive.ics.uci.edu/ml/datasets/Adult}. The target variable indicates whether a person earns more then 50k dollars per year. The goal is to predict the target feature based on census data. In this dataset, we considered as sensitive features: ``Marital Status”, ``Race”, and ``Sex”.

\textbf{German.} This is also a dataset available on UCI repository\footnote{https://archive.ics.uci.edu/ml/datasets/statlog+(german+credit+data)}. The task is to predict if an applicant has a high credit risk. In other words, if an applicant is likely to pay back his loan. We considered as sensitive features: ``statussex'', ``telephone", and ``foreign worker".

\textbf{HMDA.}  The {\it Home Mortgage Disclosure Act} (HMDA)\footnote{https://www.consumerfinance.gov/data-research/hmda/} aims to help identifying possible discriminatory lending practices. This public data about home mortgage  contains information about the applicant (demographic information), the lender (name, regulator), the property (type of property, owner occupancy, census tract), and the loan (loan amount, type of loan, loan purpose). Here, the goal is to predict whether a loan is ``high-priced", and the features that are considered sensitive are ``sex", ``race", and ``ethnicity".

\textbf{Default.} This dataset is also a dataset available on the UCI repository\footnote{https://archive.ics.uci.edu/ml/datasets/default+of+credit+card+clients}.
The goal is to predict the probability of default payments using data from Taiwanese credit card users, e.g., credit limit, gender, education, marital status, history of payment, bill and payment amounts. We consider as sensitive features in this dataset: ``sex'' and ``marriage''.

\textbf{LSAC.} The {\it Law School Admissions Council }(LSAC)\footnote{http://www.seaphe.org/databases.php} dataset contains information about approx. 27K  students through law school, graduation, and sittings for bar exams. This information was collected from 1991 through 1997, and it describes students' gender, race, year of birth (\texttt{DOB\_yr}), full-time status, family income, Law School Admission Test score (\texttt{lsat}), and academic performace (undegraduate GPA (\texttt{ugpa}), standardized overall GPA (\texttt{zgpa}), standardized 1st year GPA (\texttt{zfygpa}),  weighted index using 60\% of LSAT and 40\% of \texttt{ugpa} (\texttt{weighted\_lsat\_ugpa})). Here, the goal is to predict whether a law student passes in the bar exam. In this dataset, features that could be considered sensitive are ``race" and ``sex".

\subsection{Empirical Setup}\label{ssec:empericalset}

To perform our experiments\footnote{The gitlab repository of LimeOut can be found here: \url{https://gitlab.inria.fr/orpailleur/limeout}}, we split each dataset into 70\% training set and 30\% testing. As the datasets are imbalanced, we used Synthetic Minority Oversampling Technique (SMOTE\footnote{https://machinelearningmastery.com/threshold-moving-for-imbalanced-classification/}) over training data to generate the samples synthetically. 
We trained original and ensemble models on the balanced (augmented) datasets using Scikit-learn implementations~\cite{scikit-learn} of the following five algorithms:  \textit{AdaBoost} (ADA), \textit{Bagging}, \textit{Random Forest} (RF), and \textit{Logistic Regression} (LR).
For ADA, Bagging, RF, and LR we kept the default parameters of Scikit-learn documentation\footnote{We used version 0.23.1 of Scikit-learn.}.

 \subsection{Accuracy Assessment}\label{sec:acc-assessment}
 Table \ref{tb:accuracy} shows the average accuracy obtained in all experiments. We repeated the same experiment 10 times. For each dataset, we indicate the average accuracy of the original model (``Original'') and the average accuracy of the \F \,ensemble model (line ``\F''). Our analysis is based on the comparison between the accuracy of the original and the ensemble models. Since we drop sensitive features, it is expected that the accuracy of model decreases. However, it is evident that \F \,ensemble models maintain the level of accuracy, even though sensitive features were dropped out. 
\\
 We notice a slight improvement in the accuracy of the ensemble models when we use Bagging over German, Adult and Default datasets. 
 Although in some cases we notice a difference between original and ensemble models, in all scenarios the difference is statistically negligible.

 \begin{table*}
 \caption{Average accuracy assessment, where \F \,stands for the ensemble model built by our proposed framework. Numbers in parentheses indicate standard deviation. No accuracy values are reported on the HMDA dataset for logistic regression, and on the Default dataset for random forest and logistic regression, since in each of these cases the original model was deemed fair.}
 \centering
 \begin{tabular}{r|r|llll}
 \hline
         &          & ADA      & Bagging       & RF            & LR               \\
 \hline
 German  & Original & 0.757 (0.015) & 0.743 (0.019) & \textbf{0.772} (0.016) & {0.769} (0.021)  \\
         & \F  \,     & 0.765 (0.014) & 0.755 (0.021) & 0.769 (0.016) & 0.770 (0.021)  \\
         \hline
 Adult   & Original & \textbf{0.855} (0.003) & 0.841 (0.002) & 0.808 (0.007) & 0.845 (0.004)  \\
         & \F   \,    & 0.856 (0.003) & 0.849 (0.002) & 0.808 (0.004) & 0.849 (0.004)  \\
         \hline
 HMDA    & Original & 0.879 (0.001) & \textbf{0.883} (0.001) & 0.882 (0.001) & 0.878 (0.001)               \\
         & \F    \,   & 0.880 (0.001) & 0.884 (0.000) & 0.884 (0.000) & -                       \\
         \hline
 LSAC    & Original & 0.857 (0.003) & \textbf{0.861} (0.002) & 0.852 (0.002) & 0.820 (0.006)  \\
         & \F   \,    & 0.859 (0.002) & 0.866 (0.002) & 0.859 (0.002) & 0.822 (0.005)  \\
         \hline
 Default & Original & \textbf{0.817} (0.003) & 0.804 (0.003) & 0.807 (0.003) & 0.779 (0.004)            \\
         & \F    \,   & 0.817 (0.003) & 0.812 (0.002) & -             & -                    \\
         \hline
 \end{tabular}
 
 \label{tb:accuracy}
 \end{table*}

\subsection{Fairness Assessment}\label{sec:exp-assessment}

We now assess model fairness with respect to two points of view, namely, in terms of {\it process fairness}  and in terms of various {\it fairness metrics}.

\subsubsection{Process Fairness.} 

In this section we analyze the impact of feature dropout and the dependence on sensitive features. We employ LIME\textsubscript{Global} to compute feature contributions and build the list of the most important features. Instead of providing the lists of feature contributions for all combinations of datasets and classifiers, for each dataset, we select the classifier that provides the highest accuracy, as we did in Subsection \ref{sec:acc-assessment}.
\\
We thus look at the explanations obtained from LIME\textsubscript{Global}
for these selected combinations. Tables \ref{tb:exp-adult}, \ref{tb:exp-german}, \ref{tb:exp-hmda}, \ref{tb:exp-default} and \ref{tb:exp-lsac} present the list of most important features for these datasets. In all cases, we can notice that \F decreases the dependence on sensitive features. In other words, the ensemble models provided by our framework have less sensitive features in the list of most important features. Also, LIME explanations show that the remaining sensitive features (the ones that appeared in the list of the ensemble model) contributed less to the global prediction compared to the original model. 
 \\
For all datasets we used $k=10$, except for the HMDA dataset. Indeed, in the latter case we took $k=15$ (Table \ref{tb:exp-hmda}). This is due to the fact that all models were considered fair by  \F if only the first 10 important features were taken into account. We thus decided to investigate whether considering more features would show a different result, as it turned out to be the case when applying Bagging on HMDA.  

\begin{table*}[!htb]
    \caption{LIME explanations in the form of pairs feature/contribution for the original AdaBoost model and the ensemble variant (\F's output) on the  on Adult dataset. }
    \label{tb:exp-adult}
    \footnotesize
    \begin{subtable}{0.5\linewidth}
        \centering
        \setlength{\tabcolsep}{1.9pt}
        \begin{tabular}{ll}
            \multicolumn{2}{c}{Original}   \\
            \hline
            \textbf{Feature} & \textbf{Contrib.} \\
            \hline
            CapitalGain & -18.449067 \\ 
            CapitalLoss & -4.922207 \\ 
            Hoursperweek & 3.297749 \\ 
            Workclass & -0.997601 \\ 
            fnlwgt & -0.890244 \\ 
            \textbf{MaritalStatus} & 0.873829 \\ 
            \textbf{Sex} & 0.694676 \\ 
            Education-Num & -0.603877 \\ 
            Relationship & 0.277705 \\ 
            Occupation & 0.173059 \\ 
            \hline
        \end{tabular}
    \end{subtable}%
    \begin{subtable}{0.5\linewidth}
        \centering
        \setlength{\tabcolsep}{1.9pt}
        \begin{tabular}{ll}
            \multicolumn{2}{c}{Ensemble}   \\
            \hline
            \textbf{Feature} & \textbf{Contrib.} \\
            \hline
            CapitalGain & -19.147673 \\ 
            CapitalLoss & -9.682837 \\ 
            Hoursperweek & 1.173417 \\ 
            fnlwgt & 0.974685 \\ 
            Workclass & -0.423646 \\ 
            Education-Num & -0.259837 \\ 
            \textbf{Sex} & -0.244728 \\ 
            Country & -0.162728 \\ 
            Education & 0.127105 \\ 
            Age & -0.124858 \\             
            \hline
    \end{tabular}
    \end{subtable} 
\end{table*}

\begin{table*}[!htb]
    \caption{LIME explanations of RF on German dataset.}
    \label{tb:exp-german}
    \footnotesize
    \begin{subtable}{0.5\linewidth}
        \centering
        \setlength{\tabcolsep}{1.9pt}
        \begin{tabular}{ll}
            \multicolumn{2}{c}{Original}   \\
            \hline
            \textbf{Feature} & \textbf{Contrib.} \\
            \hline
            \textbf{foreignworker} & 2.664899 \\ 
            otherinstallmentplans & -1.354191 \\ 
            housing & -1.144371 \\ 
            savings & 0.984104 \\ 
            property & -0.648104 \\ 
            purpose & -0.415498 \\ 
            existingchecking & 0.371415 \\ 
            \textbf{telephone} & 0.311451 \\ 
            credithistory & 0.263366 \\ 
            duration & -0.223288 \\ 
            \hline
        \end{tabular}
    \end{subtable}%
    \begin{subtable}{0.5\linewidth}
        \centering
        \setlength{\tabcolsep}{1.9pt}
        \begin{tabular}{ll}
            \multicolumn{2}{c}{Ensemble}   \\
            \hline
             \textbf{Feature} & \textbf{Contrib.} \\
            \hline
            otherinstallmentplans & -1.487604 \\ 
            housing & -1.089726 \\ 
            savings & 0.679195 \\ 
            duration & -0.483643 \\ 
            \textbf{foreignworker} & 0.448643 \\ 
            property & -0.386355 \\ 
            credithistory & 0.258375 \\ 
            job & -0.252046 \\ 
            existingchecking & -0.21358 \\ 
            residencesince & -0.138818 \\            
            \hline
    \end{tabular}
    \end{subtable} 
    
\end{table*}

\begin{table*}[!htb]
    \caption{LIME explanation of Bagging on HMDA dataset.}
    \label{tb:exp-hmda}
    \footnotesize
    \begin{subtable}{0.5\linewidth}
        \centering
        \setlength{\tabcolsep}{1.9pt}
        \begin{tabular}{ll}
            \multicolumn{2}{c}{Original}   \\
            \hline
            \textbf{Feature} & \textbf{Contrib.} \\
            \hline
            derived\_loan\_product\_type & 4.798847 \\ 
            balloon\_payment\_desc & 4.624029 \\ 
            intro\_rate\_period & 4.183828 \\ 
            loan\_to\_value\_ratio & 2.824717 \\ 
            balloon\_payment & 2.005847 \\ 
            prepayment\_penalty\_term & 0.683618 \\ 
            reverse\_mortgage & -0.659169 \\ 
            applicant\_age\_above\_62 & 0.532331 \\ 
            \textbf{derived\_ethnicity} & -0.409255 \\ 
            co\_applicant\_age\_above\_62 & -0.333838 \\ 
            property\_value & -0.326801 \\ 
            \textbf{derived\_race} & -0.318802 \\ 
            applicant\_age & -0.304565 \\ 
            loan\_term & 0.270951 \\ 
            negative\_amortization & -0.229379 \\
            \hline
        \end{tabular}
    \end{subtable}%
    \begin{subtable}{0.5\linewidth}
        \centering
        \setlength{\tabcolsep}{1.9pt}
        \begin{tabular}{ll}
            \multicolumn{2}{c}{Ensemble}   \\
            \hline
            \textbf{Feature} & \textbf{Contrib.} \\
            \hline
            derived\_loan\_product\_type & 6.457707 \\ 
            balloon\_payment\_desc & 5.054243 \\ 
            intro\_rate\_period & 4.638744 \\ 
            balloon\_payment & 1.512304 \\ 
            prepayment\_penalty\_term & -1.267424 \\ 
            interest\_only\_payment & 0.777766 \\ 
            loan\_to\_value\_ratio & 0.704758 \\ 
            negative\_amortization\_desc & 0.61936 \\ 
            reverse\_mortgage\_desc & 0.508204 \\ 
            interest\_only\_payment\_desc & -0.393068 \\ 
            applicant\_credit\_score\_type\_desc & -0.379852 \\ 
            negative\_amortization & -0.353717 \\ 
            applicant\_age\_above\_62 & 0.349847 \\ 
            property\_value & -0.316311 \\ 
            applicant\_credit\_score\_type & -0.192114 \\ 
            \hline
    \end{tabular}
    \end{subtable}    
\end{table*}

\begin{table*}[!htb]
    \caption{LIME explanations of AdaBoost on Default dataset.}
    \label{tb:exp-default}
    \footnotesize
    \begin{subtable}{0.5\linewidth}
        \centering
        \setlength{\tabcolsep}{1.9pt}
        \begin{tabular}{ll}
            \multicolumn{2}{c}{Original}   \\
            \hline
            \textbf{Feature} & \textbf{Contrib.} \\
            \hline
            PAY\_0 & 0.014194 \\ 
            \textbf{MARRIAGE} & -0.013986 \\ 
            PAY\_2 & -0.013513 \\ 
            PAY\_6 & -0.011724 \\ 
            PAY\_AMT1 & 0.011664 \\ 
            PAY\_AMT6 & 0.008088 \\ 
            PAY\_AMT2 & 0.007735 \\ 
            PAY\_3 & 0.00735 \\ 
            EDUCATION & 0.0032 \\ 
            \textbf{SEX} & 0.000732 \\ 
            \hline
        \end{tabular}
    \end{subtable}%
    \begin{subtable}{0.5\linewidth}
        \centering
        \setlength{\tabcolsep}{1.9pt}
        \begin{tabular}{ll}
            \multicolumn{2}{c}{Ensemble}   \\
            \hline
            \textbf{Feature} & \textbf{Contrib.} \\
            \hline
            PAY\_2 & -0.024354 \\ 
            PAY\_0 & 0.008862 \\ 
            PAY\_5 & 0.008729 \\ 
            PAY\_AMT6 & -0.00566 \\ 
            LIMIT\_BAL & -0.003584 \\ 
            BILL\_AMT2 & 0.00329 \\ 
            PAY\_6 & -0.00307 \\ 
            AGE & -0.002058 \\ 
            PAY\_AMT1 & 0.001592 \\ 
            PAY\_3 & -0.001492 \\ 
            \hline
    \end{tabular}
    \end{subtable}      
\end{table*}

\begin{table*}[!htb]
    \caption{LIME explanations of Bagging on LSAC dataset.}
    \label{tb:exp-lsac}
    \footnotesize
    \begin{subtable}{0.5\linewidth}
        \centering
        \setlength{\tabcolsep}{1.9pt}
        \begin{tabular}{ll}
            \multicolumn{2}{c}{Original}   \\
            \hline
            \textbf{Feature} & \textbf{Contrib.} \\
            \hline
            isPartTime & -12.588169 \\ 
            \textbf{race} & -3.943962 \\ 
            cluster\_tier & -1.873394 \\ 
            DOB\_yr & -1.235803 \\ 
            zgpa & -0.71457 \\ 
            zfygpa & 0.314865 \\ 
            ugpa & 0.123805 \\ 
            family\_income & -0.08999 \\ 
            lsat & -0.07596 \\ 
            \textbf{sex} & -0.068117 \\ 
            \hline
        \end{tabular}
    \end{subtable}%
    \begin{subtable}{0.5\linewidth}
        \centering
        \setlength{\tabcolsep}{1.9pt}
        \begin{tabular}{ll}
            \multicolumn{2}{c}{Ensemble}   \\
            \hline
            \textbf{Feature} & \textbf{Contrib.} \\
            \hline
            isPartTime & -9.294158 \\ 
            cluster\_tier & -3.464014 \\ 
            zgpa & 2.835836 \\ 
            family\_income & -1.292526 \\ 
            DOB\_yr & -0.923861 \\ 
            \textbf{race} & -0.895484 \\ 
            zfygpa & 0.238397 \\ 
            weighted\_lsat\_ugpa & 0.060846 \\ 
            ugpa & -0.055593 \\ 
            \textbf{sex} & -0.041478 \\
            \hline
    \end{tabular}
    \end{subtable} 
\end{table*}

\subsubsection{Fairness Metrics}

\begin{figure*}
\begin{center}
\begin{tikzpicture}[scale=1.0]
	\footnotesize
    \begin{customlegend}[legend entries={Original, Ensemble}, 
    legend columns=2, legend style={draw=none}]
    	\addlegendimage{only marks, mark=*,blue}
	    \addlegendimage{only marks, mark=*,red}
    \end{customlegend}
\end{tikzpicture}
\makebox[\linewidth][c]{%
\begin{subfigure}[]{0.22\textwidth}
\caption{DP}
\begin{tikzpicture}
\begin{axis}[
    height=4cm,
 	width=4.5cm,
	ymax=0.4,
	ymin=-0.4,
 	xticklabel style = {font=\tiny,yshift=0ex},
 	yticklabel style = {font=\tiny},
 	x tick label style={at={(0.5,-0.25)},rotate=50,anchor=east},
 	xlabel style={yshift=-50pt},
 	ylabel style={yshift=-15pt},
    scatter/classes={original={mark=*,blue,mark size=1.5}, ensemble={mark=*,red,mark size=1.5}},
	symbolic x coords={ADA-sex,ADA-marital,ADA-race,BAGGING-marital,BAGGING-sex,RF-sex,RF-marital,RF-race,LR-sex,LR-marital,LR-race},
	xtick=data
	]
	
    
    \coordinate (A) at (axis cs:BAGGING-marital,0);
    \coordinate (O1) at (rel axis cs:0,0);
    \coordinate (O2) at (rel axis cs:1,0);
    
    \draw [black,dashed] (A -| O1) -- (A -| O2);
    
	\addplot[scatter,only marks,scatter src=explicit symbolic] coordinates {
(LR-sex,-0.08398062)[original]
(LR-marital,0.069778192)[original]
(LR-race,0.024622262)[original]
(LR-sex,-0.04727434)[ensemble]
(LR-marital,0.027380726)[ensemble]
(LR-race,0.024622262)[ensemble]
(ADA-marital,0.16421011)[original]
(ADA-sex,-0.184724169)[original]
(ADA-race,0.079328473)[original]
(ADA-marital,0.166752068)[ensemble]
(ADA-sex,-0.178731892)[ensemble]
(ADA-race,0.084776379)[ensemble]
(RF-sex,-0.199657032)[original]
(RF-marital,0.193570162)[original]
(RF-race,0.152011249)[original]
(RF-sex,-0.188522765)[ensemble]
(RF-marital,0.178452478)[ensemble]
(RF-race,0.139514513)[ensemble]
(BAGGING-marital,0.141443389)[original]
(BAGGING-sex,-0.154187397)[original]
(BAGGING-marital,0.158970136)[ensemble]
(BAGGING-sex,-0.166509169)[ensemble]

};
\end{axis}
\end{tikzpicture}
\vspace*{-5mm}
\end{subfigure}
\begin{subfigure}[]{0.22\textwidth}
\caption{EQ}
\begin{tikzpicture}
\begin{axis}[
    height=4cm,
 	width=4.5cm,
	ymax=0.4,
	ymin=-0.4,
 	xticklabel style = {font=\tiny,yshift=0ex},
 	yticklabel style = {font=\tiny},
 	x tick label style={at={(0.5,-0.25)},rotate=50,anchor=east},
 	xlabel style={yshift=-50pt},
 	ylabel style={yshift=-15pt},
    scatter/classes={original={mark=*,blue,mark size=1.5}, ensemble={mark=*,red,mark size=1.5}},
	symbolic x coords={ADA-sex,ADA-marital,ADA-race,BAGGING-marital,BAGGING-sex,RF-sex,RF-marital,RF-race,LR-sex,LR-marital,LR-race},
	xtick=data
	]
	
    
    \coordinate (A) at (axis cs:BAGGING-marital,0);
    \coordinate (O1) at (rel axis cs:0,0);
    \coordinate (O2) at (rel axis cs:1,0);
    
    \draw [black,dashed] (A -| O1) -- (A -| O2);
    
	\addplot[scatter,only marks,scatter src=explicit symbolic] coordinates {

(LR-sex,-0.041091387)[original]
(LR-marital,0.039673339)[original]
(LR-race,-0.009434172)[original]
(LR-sex,0.022250759)[ensemble]
(LR-marital,-0.075400302)[ensemble]
(LR-race,-0.009434172)[ensemble]
(ADA-marital,0.285042437)[original]
(ADA-sex,-0.165299081)[original]
(ADA-race,0.025995592)[original]
(ADA-marital,0.303918387)[ensemble]
(ADA-sex,-0.145773919)[ensemble]
(ADA-race,0.041739887)[ensemble]
(RF-sex,-0.145672526)[original]
(RF-marital,0.349046877)[original]
(RF-race,0.296344132)[original]
(RF-sex,-0.134761624)[ensemble]
(RF-marital,0.304964963)[ensemble]
(RF-race,0.227534347)[ensemble]
(BAGGING-marital,0.194674285)[original]
(BAGGING-sex,-0.092682966)[original]
(BAGGING-marital,0.222815892)[ensemble]
(BAGGING-sex,-0.085881007)[ensemble]

};
\end{axis}
\end{tikzpicture}
\vspace*{-5mm}
\end{subfigure}
\begin{subfigure}[]{0.22\textwidth}
\caption{EQ}
\begin{tikzpicture}
\begin{axis}[
    height=4cm,
 	width=4.5cm,
	ymax=0.4,
	ymin=-0.4,
 	xticklabel style = {font=\tiny,yshift=0ex},
 	yticklabel style = {font=\tiny},
 	x tick label style={at={(0.5,-0.25)},rotate=50,anchor=east},
 	xlabel style={yshift=-50pt},
 	ylabel style={yshift=-15pt},
    scatter/classes={original={mark=*,blue,mark size=1.5}, ensemble={mark=*,red,mark size=1.5}},
	symbolic x coords={ADA-sex,ADA-marital,ADA-race,BAGGING-marital,BAGGING-sex,RF-sex,RF-marital,RF-race,LR-sex,LR-marital,LR-race},
	xtick=data
	]
	
    
    \coordinate (A) at (axis cs:BAGGING-marital,0);
    \coordinate (O1) at (rel axis cs:0,0);
    \coordinate (O2) at (rel axis cs:1,0);
    
    \draw [black,dashed] (A -| O1) -- (A -| O2);
    
	\addplot[scatter,only marks,scatter src=explicit symbolic] coordinates {

(LR-sex,0.145425121)[original]
(LR-marital,-0.114863399)[original]
(LR-race,-0.076340536)[original]
(LR-sex,0.150326074)[ensemble]
(LR-marital,-0.121160424)[ensemble]
(LR-race,-0.076340536)[ensemble]
(ADA-marital,-0.070230625)[original]
(ADA-sex,0.105061638)[original]
(ADA-race,-0.05363446)[original]
(ADA-marital,-0.068063784)[ensemble]
(ADA-sex,0.104015264)[ensemble]
(ADA-race,-0.05357099)[ensemble]
(RF-sex,0.114799076)[original]
(RF-marital,-0.076426281)[original]
(RF-race,-0.054331997)[original]
(RF-sex,0.112880747)[ensemble]
(RF-marital,-0.076983219)[ensemble]
(RF-race,-0.059762928)[ensemble]
(BAGGING-marital,-0.087937597)[original]
(BAGGING-sex,0.118867449)[original]
(BAGGING-marital,-0.085398476)[ensemble]
(BAGGING-sex,0.115435122)[ensemble]

};
\end{axis}
\end{tikzpicture}
\vspace*{-5mm}
\end{subfigure}
\begin{subfigure}[]{0.22\textwidth}
\caption{PE}
\begin{tikzpicture}
\begin{axis}[
    height=4cm,
 	width=4.5cm,
	ymax=0.4,
	ymin=-0.4,
 	xticklabel style = {font=\tiny,yshift=0ex},
 	yticklabel style = {font=\tiny},
 	x tick label style={at={(0.5,-0.25)},rotate=50,anchor=east},
 	xlabel style={yshift=-50pt},
 	ylabel style={yshift=-20pt},
    scatter/classes={original={mark=*,blue,mark size=1.5}, ensemble={mark=*,red,mark size=1.5}},
	symbolic x coords={ADA-sex,ADA-marital,ADA-race,BAGGING-marital,BAGGING-sex,RF-sex,RF-marital,RF-race,LR-sex,LR-marital,LR-race},
	xtick=data,
	legend pos=outer north east,
	legend cell align=left
	]
	
    
    \coordinate (A) at (axis cs:BAGGING-marital,0);
    \coordinate (O1) at (rel axis cs:0,0);
    \coordinate (O2) at (rel axis cs:1,0);
    
    \draw [black,dashed] (A -| O1) -- (A -| O2);
    
	\addplot[scatter,only marks,scatter src=explicit symbolic] coordinates {

(LR-sex,-0.027918501)[original]
(LR-marital,0.023483257)[original]
(LR-race,-0.000127076)[original]
(LR-sex,-0.004854142)[ensemble]
(LR-marital,-0.000411642)[ensemble]
(LR-race,-0.000127076)[ensemble]
(ADA-marital,0.057926204)[original]
(ADA-sex,-0.070515889)[original]
(ADA-race,0.010162825)[original]
(ADA-marital,0.057938783)[ensemble]
(ADA-sex,-0.065921781)[ensemble]
(ADA-race,0.013724835)[ensemble]
(RF-sex,-0.090966189)[original]
(RF-marital,0.081960074)[original]
(RF-race,0.059671998)[original]
(RF-sex,-0.080516398)[ensemble]
(RF-marital,0.071770942)[ensemble]
(RF-race,0.054510747)[ensemble]
(BAGGING-marital,0.054611287)[original]
(BAGGING-sex,-0.060462968)[original]
(BAGGING-marital,0.064065682)[ensemble]
(BAGGING-sex,-0.067030096)[ensemble]

};

\end{axis}
\end{tikzpicture}
\vspace*{-5mm}
\end{subfigure}
\begin{subfigure}[]{0.22\textwidth}
\caption{DI}
\begin{tikzpicture}
\begin{axis}[
    height=4cm,
 	width=4.5cm,
	ymax=6,
	ymin=-4,
 	xticklabel style = {font=\tiny,yshift=0ex},
 	yticklabel style = {font=\tiny},
 	x tick label style={at={(0.5,-0.25)},rotate=50,anchor=east},
 	xlabel style={yshift=-50pt},
 	ylabel style={yshift=-15pt},
    scatter/classes={original={mark=*,blue,mark size=1.5}, ensemble={mark=*,red,mark size=1.5}},
	symbolic x coords={ADA-sex,ADA-marital,ADA-race,BAGGING-marital,BAGGING-sex,RF-sex,RF-marital,RF-race,LR-sex,LR-marital,LR-race},
	xtick=data,
	legend pos=outer north east,
	legend cell align=left
	]
	
    
    \coordinate (A) at (axis cs:LR-sex,1);
    \coordinate (O1) at (rel axis cs:0,0);
    \coordinate (O2) at (rel axis cs:1,0);
    
    \draw [black,dashed] (A -| O1) -- (A -| O2);
    
	\addplot[scatter,only marks,scatter src=explicit symbolic] coordinates {
(LR-sex,0.383520142)[original]
(LR-marital,2.778185547)[original]
(LR-race,2.614589062)[original]
(LR-sex,0.474717903)[ensemble]
(LR-marital,1.553410138)[ensemble]
(LR-race,2.614589062)[ensemble]
(ADA-marital,4.753111253)[original]
(ADA-sex,0.254054372)[original]
(ADA-race,1.866879597)[original]
(ADA-marital,5.089393395)[ensemble]
(ADA-sex,0.270636631)[ensemble]
(ADA-race,1.895553207)[ensemble]
(RF-sex,0.27431854)[original]
(RF-marital,5.669890678)[original]
(RF-race,4.636656267)[original]
(RF-sex,0.275102825)[ensemble]
(RF-marital,5.141311281)[ensemble]
(RF-race,3.904397211)[ensemble]
(BAGGING-marital,3.883444969)[original]
(BAGGING-sex,0.307441428)[original]
(BAGGING-marital,4.294934105)[ensemble]
(BAGGING-sex,0.308424911)[ensemble]

};
\end{axis}
\end{tikzpicture}
\vspace*{-5mm}
\end{subfigure}
}
\end{center}
\caption{Fairness metrics for Adult Dataset.}
\label{fig:adult_metrics}
\end{figure*}
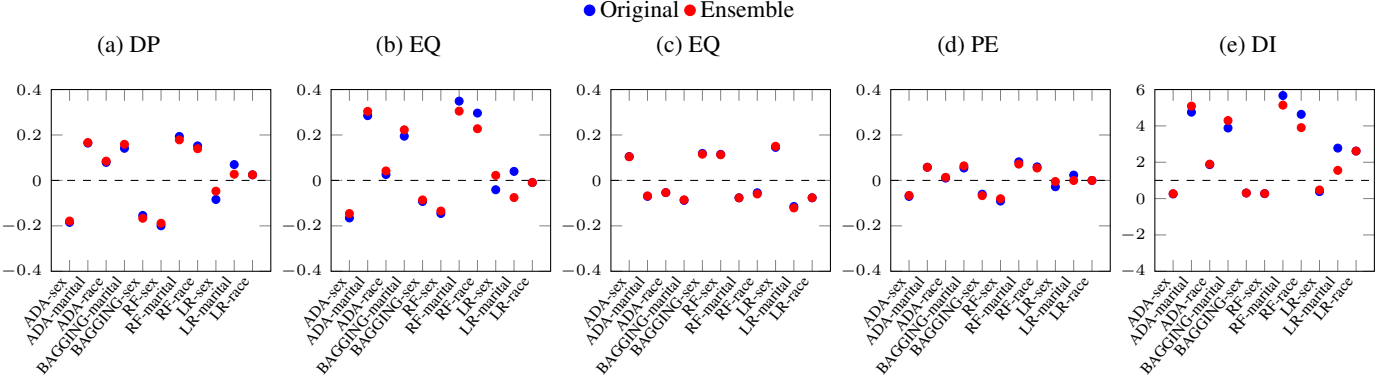

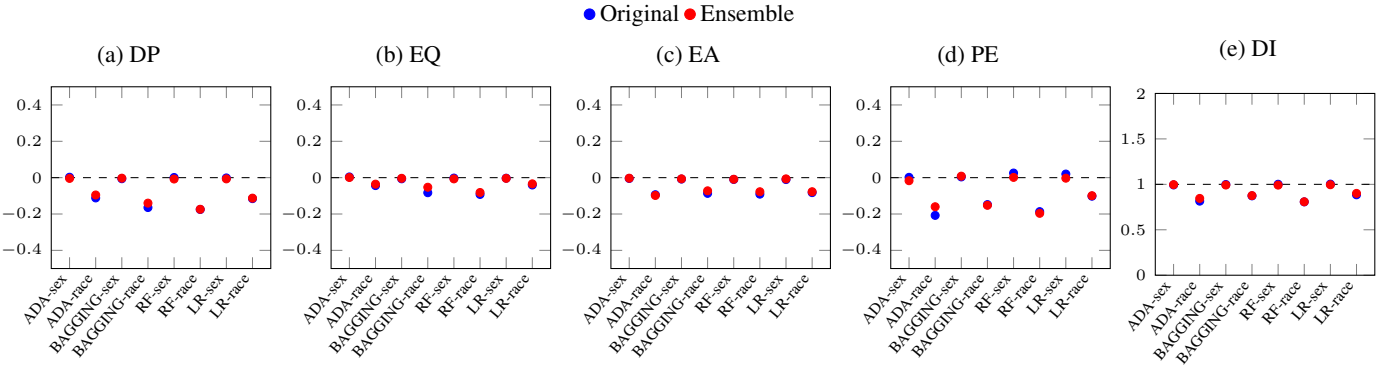
\begin{figure*}
\begin{center}
\begin{tikzpicture}[scale=1.0]
	\footnotesize
    \begin{customlegend}[legend entries={Original, Ensemble}, 
    legend columns=2, legend style={draw=none}]
    	\addlegendimage{only marks, mark=*,blue}
	    \addlegendimage{only marks, mark=*,red}
    \end{customlegend}
\end{tikzpicture}
\makebox[\linewidth][c]{%
\begin{subfigure}[]{0.22\textwidth}
\caption{DP}
\begin{tikzpicture}
\begin{axis}[
    height=4cm,
 	width=4.5cm,
	ymax=0.5,
	ymin=-0.5,
 	xticklabel style = {font=\tiny,yshift=0ex},
 	yticklabel style = {font=\tiny},
 	x tick label style={at={(0.5,-0.25)},rotate=50,anchor=east},
 	xlabel style={yshift=-50pt},
 	ylabel style={yshift=-20pt},
    scatter/classes={original={mark=*,blue,mark size=1.5}, ensemble={mark=*,red,mark size=1.5}},
	symbolic x coords={ADA-sex,ADA-race,BAGGING-sex,BAGGING-race,RF-sex,RF-race,LR-sex,LR-race},
	xtick=data,
	legend pos=outer north east,
	legend cell align=left
	]
	
    
    \coordinate (A) at (axis cs:ADA-sex,0);
    \coordinate (O1) at (rel axis cs:0,0);
    \coordinate (O2) at (rel axis cs:1,0);
    
    \draw [black,dashed] (A -| O1) -- (A -| O2);
    
	\addplot[scatter,only marks,scatter src=explicit symbolic] coordinates {

(ADA-race,-0.111596625)[original]
(ADA-sex,0.002370738)[original]
(ADA-race,-0.095358436)[ensemble]
(ADA-sex,-0.004161645)[ensemble]
(BAGGING-race,-0.164477569)[original]
(BAGGING-sex,-0.00594963)[original]
(BAGGING-race,-0.140173266)[ensemble]
(BAGGING-sex,-0.002863843)[ensemble]
(LR-race,-0.115525983)[original]
(LR-sex,-0.001498329)[original]
(LR-race,-0.11272715)[ensemble]
(LR-sex,-0.006792697)[ensemble]
(RF-race,-0.175282725)[original]
(RF-sex,0.000935689)[original]
(RF-race,-0.173957778)[ensemble]
(RF-sex,-0.00751972)[ensemble]


};

\end{axis}
\end{tikzpicture}
\vspace*{-5mm}
\end{subfigure}
\begin{subfigure}[]{0.22\textwidth}
\caption{EQ}
\begin{tikzpicture}
\begin{axis}[
    height=4cm,
 	width=4.5cm,
	ymax=0.5,
	ymin=-0.5,
 	xticklabel style = {font=\tiny,yshift=0ex},
 	yticklabel style = {font=\tiny},
 	x tick label style={at={(0.5,-0.25)},rotate=50,anchor=east},
 	xlabel style={yshift=-50pt},
 	ylabel style={yshift=-20pt},
    scatter/classes={original={mark=*,blue,mark size=1.5}, ensemble={mark=*,red,mark size=1.5}},
	symbolic x coords={ADA-sex,ADA-race,BAGGING-sex,BAGGING-race,RF-sex,RF-race,LR-sex,LR-race},
	xtick=data,
	legend pos=outer north east,
	legend cell align=left
	]
	
    
    \coordinate (A) at (axis cs:ADA-sex,0);
    \coordinate (O1) at (rel axis cs:0,0);
    \coordinate (O2) at (rel axis cs:1,0);
    
    \draw [black,dashed] (A -| O1) -- (A -| O2);
    
	\addplot[scatter,only marks,scatter src=explicit symbolic] coordinates {
(ADA-race,-0.043678051)[original]
(ADA-sex,0.004039658)[original]
(ADA-race,-0.035821128)[ensemble]
(ADA-sex,0.00084871)[ensemble]
(BAGGING-race,-0.082873997)[original]
(BAGGING-sex,-0.006421752)[original]
(BAGGING-race,-0.05273932)[ensemble]
(BAGGING-sex,-0.00360854)[ensemble]
(LR-race,-0.040383559)[original]
(LR-sex,-0.003004606)[original]
(LR-race,-0.034022785)[ensemble]
(LR-sex,-0.003844859)[ensemble]
(RF-race,-0.092410627)[original]
(RF-sex,-0.002258585)[original]
(RF-race,-0.081588308)[ensemble]
(RF-sex,-0.00678268)[ensemble]


};

\end{axis}
\end{tikzpicture}
\vspace*{-5mm}
\end{subfigure}
\begin{subfigure}[]{0.22\textwidth}
\caption{EA}
\begin{tikzpicture}
\begin{axis}[
    height=4cm,
 	width=4.5cm,
	ymax=0.5,
	ymin=-0.5,
 	xticklabel style = {font=\tiny,yshift=0ex},
 	yticklabel style = {font=\tiny},
 	x tick label style={at={(0.5,-0.25)},rotate=50,anchor=east},
 	xlabel style={yshift=-50pt},
 	ylabel style={yshift=-20pt},
    scatter/classes={original={mark=*,blue,mark size=1.5}, ensemble={mark=*,red,mark size=1.5}},
	symbolic x coords={ADA-sex,ADA-race,BAGGING-sex,BAGGING-race,RF-sex,RF-race,LR-sex,LR-race},
	xtick=data,
	legend pos=outer north east,
	legend cell align=left
	]
	
    
    \coordinate (A) at (axis cs:ADA-sex,0);
    \coordinate (O1) at (rel axis cs:0,0);
    \coordinate (O2) at (rel axis cs:1,0);
    
    \draw [black,dashed] (A -| O1) -- (A -| O2);
    
	\addplot[scatter,only marks,scatter src=explicit symbolic] coordinates {

(ADA-race,-0.093908971)[original]
(ADA-sex,-0.00448846)[original]
(ADA-race,-0.098009351)[ensemble]
(ADA-sex,-0.00293273)[ensemble]
(BAGGING-race,-0.086491505)[original]
(BAGGING-sex,-0.007894839)[original]
(BAGGING-race,-0.072410911)[ensemble]
(BAGGING-sex,-0.006488998)[ensemble]
(LR-race,-0.082318621)[original]
(LR-sex,-0.010527241)[original]
(LR-race,-0.077364515)[ensemble]
(LR-sex,-0.006625919)[ensemble]
(RF-race,-0.090802196)[original]
(RF-sex,-0.009936732)[original]
(RF-race,-0.077488235)[ensemble]
(RF-sex,-0.00875979)[ensemble]


};

\end{axis}
\end{tikzpicture}
\vspace*{-5mm}
\end{subfigure}
\begin{subfigure}[]{0.22\textwidth}
\caption{PE}
\begin{tikzpicture}
\begin{axis}[
    height=4cm,
 	width=4.5cm,
	ymax=0.5,
	ymin=-0.5,
 	xticklabel style = {font=\tiny,yshift=0ex},
 	yticklabel style = {font=\tiny},
 	x tick label style={at={(0.5,-0.25)},rotate=50,anchor=east},
 	xlabel style={yshift=-50pt},
 	ylabel style={yshift=-20pt},
    scatter/classes={original={mark=*,blue,mark size=1.5}, ensemble={mark=*,red,mark size=1.5}},
	symbolic x coords={ADA-sex,ADA-race,BAGGING-sex,BAGGING-race,RF-sex,RF-race,LR-sex,LR-race},
	xtick=data,
	legend pos=outer north east,
	legend cell align=left
	]
	
    
    \coordinate (A) at (axis cs:ADA-sex,0);
    \coordinate (O1) at (rel axis cs:0,0);
    \coordinate (O2) at (rel axis cs:1,0);
    
    \draw [black,dashed] (A -| O1) -- (A -| O2);
    
	\addplot[scatter,only marks,scatter src=explicit symbolic] coordinates {
(ADA-race,-0.207653739)[original]
(ADA-sex,0.001757274)[original]
(ADA-race,-0.160037673)[ensemble]
(ADA-sex,-0.016754731)[ensemble]
(BAGGING-race,-0.148244133)[original]
(BAGGING-sex,0.00399728)[original]
(BAGGING-race,-0.152260396)[ensemble]
(BAGGING-sex,0.008198002)[ensemble]
(LR-race,-0.102079874)[original]
(LR-sex,0.019833298)[original]
(LR-race,-0.099176464)[ensemble]
(LR-sex,-0.0029268)[ensemble]
(RF-race,-0.186934192)[original]
(RF-sex,0.025354528)[original]
(RF-race,-0.196310125)[ensemble]
(RF-sex,0.001380126)[ensemble]


};

\end{axis}
\end{tikzpicture}
\vspace*{-5mm}
\end{subfigure}
\begin{subfigure}[]{0.22\textwidth}
\caption{DI}
\begin{tikzpicture}
\begin{axis}[
    height=4cm,
 	width=4.5cm,
	ymax=2,
	ymin=-0,
 	xticklabel style = {font=\tiny,yshift=0ex},
 	yticklabel style = {font=\tiny},
 	x tick label style={at={(0.5,-0.25)},rotate=50,anchor=east},
 	xlabel style={yshift=-50pt},
 	ylabel style={yshift=-15pt},
    scatter/classes={original={mark=*,blue,mark size=1.5}, ensemble={mark=*,red,mark size=1.5}},
	symbolic x coords={ADA-sex,ADA-race,BAGGING-sex,BAGGING-race,RF-sex,RF-race,LR-sex,LR-race},
	xtick=data,
	legend pos=outer north east,
	legend cell align=left
	]
	
    
    \coordinate (A) at (axis cs:ADA-sex,1);
    \coordinate (O1) at (rel axis cs:0,0);
    \coordinate (O2) at (rel axis cs:1,0);
    
    \draw [black,dashed] (A -| O1) -- (A -| O2);
    
	\addplot[scatter,only marks,scatter src=explicit symbolic] coordinates {
(ADA-race,0.815163157)[original]
(ADA-sex,0.993378181)[original]
(ADA-race,0.84522724)[ensemble]
(ADA-sex,0.996759046)[ensemble]
(BAGGING-race,0.87323322)[original]
(BAGGING-sex,0.998333234)[original]
(BAGGING-race,0.876379004)[ensemble]
(BAGGING-sex,0.992399468)[ensemble]
(LR-race,0.88484241)[original]
(LR-sex,1.002600326)[original]
(LR-race,0.900573978)[ensemble]
(LR-sex,0.995666516)[ensemble]
(RF-race,0.807178746)[original]
(RF-sex,1.001095068)[original]
(RF-race,0.808130258)[ensemble]
(RF-sex,0.991474148)[ensemble]


};
\end{axis}
\end{tikzpicture}
\vspace*{-5mm}
\end{subfigure}
}
\end{center}
\caption{Fairness metrics for LSAC Dataset.}
\label{fig:lsac_metrics}
\end{figure*}

In this section, we assess fairness using the fairness metrics introduced in Section \ref{sec:related-work}. We compute fairness metrics using IBM AI Fairness 360 Toolkit\footnote{https://github.com/Trusted-AI/AIF360} \cite{aif360-oct-2018}. Our goal is to have a different perspective on the fairness of \F ensemble models since we only assessed fairness by using LIME explanations. In this analysis, we compare the original and ensemble models for each combination of classifier and sensitive feature.   
\\
Figures \ref{fig:german_metrics}, \ref{fig:adult_metrics} and \ref{fig:lsac_metrics} show values for all fairness metrics in each graphic. Red points indicate the values for \F ensemble models while blue points indicate values for original models. The dashed line is the reference for a fair model (optimal value), i.e., 0 for all metrics except DI where the optimal is 1. 
\\
Results for the German dataset are depicted in Figure \ref{fig:german_metrics}. It is evident that \F produces ensemble models that are fairer according to metrics DP and EQ.  Red points are closer to zero compared to blue points, which means that \F ensemble models are fairer than pre-trained models. We can also notice general improvement on DI. However, we observe that the only problematic sensitive feature  is ``foreignworker'', where no improvement is observed. For all other sensitive features, we observe an improved fairness behaviour. In a few cases, the differences are negligible, which indicates that \F either improves or at least maintains the fairness metrics.
\\
Figure \ref{fig:adult_metrics} shows the results on fairness metrics for the Adult dataset. In this dataset, \F ensemble models keep values of all metrics in almost scenarios. We only see a deterioration of fairness when we compute EQ for Logistic Regression focuses on marital status. This behaviour means that \F at least maintain the value of fairness metrics when it reduces the dependence on sensitive features, but it cannot ensure fairness metrics closer to 0.
\\
The fairness metrics for LSAC dataset are depicted in Figure \ref{fig:lsac_metrics}. For this dataset, most of results indicate that \F maintains the fairness measurements. We can observe some exceptions, for instance, ``race'' with Bagging on PE and EQ, where an improvement is observed. This behaviour can indicate that, even if \F 's ensemble outputs are in general  less dependent on sensitive features, for some datasets  a weighted aggregation of pool classifiers should be employed (Section \ref{sssec: LimeOut}). For HMDA and Default datasets we observed a similar behaviour even though lesser classifiers were deemed unfair. The results for these two latter datasets are presented in the Appendix  \ref{sec:appendix} and the fairness metrics show a rather fair behaviour of the few models that were deemed unfair by \F.

\begin{figure*}
\begin{center}
\begin{tikzpicture}[scale=1.0]
	\footnotesize
    \begin{customlegend}[legend entries={Original, Ensemble}, 
    legend columns=2, legend style={draw=none}]
    	\addlegendimage{only marks, mark=*,blue}
	    \addlegendimage{only marks, mark=*,red}
    \end{customlegend}
\end{tikzpicture}
\makebox[\linewidth][c]{%
\centering
\begin{subfigure}[]{0.22\textwidth}
\caption{DP}
\begin{tikzpicture}
\begin{axis}[
    height=4cm,
 	width=4.5cm,
	ymax=0.4,
	ymin=-0.4,
 	xticklabel style = {font=\tiny,yshift=0ex},
 	yticklabel style = {font=\tiny},
 	x tick label style={at={(0.5,-0.25)},rotate=50,anchor=east},
 	xlabel style={yshift=-50pt},
 	ylabel style={yshift=-15pt},
    scatter/classes={original={mark=*,blue,mark size=1.5}, ensemble={mark=*,red,mark size=1.5}},
	symbolic x coords={ADA-foreign,ADA-telephone,ADA-sex,BAGGING-sex,BAGGING-foreign,BAGGING-telephone,RF-foreign,RF-telephone,RF-sex,LR-sex,LR-foreign,LR-telephone},
	xtick=data
	]
	
    
    \coordinate (A) at (axis cs:LR-sex,0);
    \coordinate (O1) at (rel axis cs:0,0);
    \coordinate (O2) at (rel axis cs:1,0);
    
    \draw [black,dashed] (A -| O1) -- (A -| O2);
    
	\addplot[scatter,only marks,scatter src=explicit symbolic] coordinates {
(LR-sex,-0.123415136)[original]
(LR-telephone,-0.112863524)[original]
(LR-foreign,0.148612806)[original]
(LR-sex,-0.034286293)[ensemble]
(LR-telephone,-0.021000326)[ensemble]
(LR-foreign,0.144811174)[ensemble]
(ADA-foreign,0.220600899)[original]
(ADA-telephone,-0.078072388)[original]
(ADA-sex,-0.089566527)[original]
(ADA-foreign,0.212645315)[ensemble]
(ADA-telephone,-0.049384392)[ensemble]
(ADA-sex,-0.035396729)[ensemble]
(RF-foreign,0.181105908)[original]
(RF-sex,-0.08588316)[original]
(RF-telephone,-0.115951372)[original]
(RF-foreign,0.119463479)[ensemble]
(RF-sex,-0.06694782)[ensemble]
(RF-telephone,-0.038681157)[ensemble]
(BAGGING-foreign,0.092483625)[original]
(BAGGING-telephone,-0.075202169)[original]
(BAGGING-sex,-0.04239952)[original]
(BAGGING-foreign,0.13778495)[ensemble]
(BAGGING-telephone,-0.023679241)[ensemble]
(BAGGING-sex,-0.036814692)[ensemble]


};
\end{axis}
\end{tikzpicture}
\vspace*{-5mm}
\end{subfigure}
\begin{subfigure}[]{0.22\textwidth}
\caption{EQ}
\begin{tikzpicture}
\begin{axis}[
    height=4cm,
 	width=4.5cm,
	ymax=0.4,
	ymin=-0.4,
 	xticklabel style = {font=\tiny,yshift=0ex},
 	yticklabel style = {font=\tiny},
 	x tick label style={at={(0.5,-0.25)},rotate=50,anchor=east},
 	xlabel style={yshift=-50pt},
 	ylabel style={yshift=-15pt},
    scatter/classes={original={mark=*,blue,mark size=1.5}, ensemble={mark=*,red,mark size=1.5}},
	symbolic x coords={ADA-foreign,ADA-telephone,ADA-sex,BAGGING-sex,BAGGING-foreign,BAGGING-telephone,RF-foreign,RF-telephone,RF-sex,LR-sex,LR-foreign,LR-telephone},
	xtick=data
	]
	
    
    \coordinate (A) at (axis cs:RF-sex,0);
    \coordinate (O1) at (rel axis cs:0,0);
    \coordinate (O2) at (rel axis cs:1,0);
    
    \draw [black,dashed] (A -| O1) -- (A -| O2);
    
	\addplot[scatter,only marks,scatter src=explicit symbolic] coordinates {
(LR-sex,-0.090022473)[original]
(LR-telephone,-0.039548089)[original]
(LR-foreign,0.065848545)[original]
(LR-sex,-0.026646622)[ensemble]
(LR-telephone,0.012775076)[ensemble]
(LR-foreign,0.050933049)[ensemble]
(ADA-foreign,0.129144353)[original]
(ADA-telephone,-0.066274388)[original]
(ADA-sex,-0.070880936)[original]
(ADA-foreign,0.110649574)[ensemble]
(ADA-telephone,-0.037974595)[ensemble]
(ADA-sex,-0.025402726)[ensemble]
(RF-foreign,0.077340376)[original]
(RF-sex,-0.058307335)[original]
(RF-telephone,-0.12338594)[original]
(RF-foreign,0.024978216)[ensemble]
(RF-sex,-0.040666369)[ensemble]
(RF-telephone,-0.040992007)[ensemble]
(BAGGING-foreign,0.015142073)[original]
(BAGGING-telephone,-0.08180778)[original]
(BAGGING-sex,-0.048569852)[original]
(BAGGING-foreign,0.039305304)[ensemble]
(BAGGING-telephone,-0.015827613)[ensemble]
(BAGGING-sex,-0.023690506)[ensemble]


};
\end{axis}
\end{tikzpicture}
\vspace*{-5mm}
\end{subfigure}
\begin{subfigure}[]{0.22\textwidth}
\caption{EA}
\begin{tikzpicture}
\begin{axis}[
    height=4cm,
 	width=4.5cm,
	ymax=0.4,
	ymin=-0.4,
 	xticklabel style = {font=\tiny,yshift=0ex},
 	yticklabel style = {font=\tiny},
 	x tick label style={at={(0.5,-0.25)},rotate=50,anchor=east},
 	xlabel style={yshift=-50pt},
 	ylabel style={yshift=-15pt},
    scatter/classes={original={mark=*,blue,mark size=1.5}, ensemble={mark=*,red,mark size=1.5}},
	symbolic x coords={ADA-foreign,ADA-telephone,ADA-sex,BAGGING-sex,BAGGING-foreign,BAGGING-telephone,RF-foreign,RF-telephone,RF-sex,LR-sex,LR-foreign,LR-telephone},
	xtick=data
	]
	
    
    \coordinate (A) at (axis cs:RF-sex,0);
    \coordinate (O1) at (rel axis cs:0,0);
    \coordinate (O2) at (rel axis cs:1,0);
    
    \draw [black,dashed] (A -| O1) -- (A -| O2);
    
	\addplot[scatter,only marks,scatter src=explicit symbolic] coordinates {
(LR-sex,-0.058560693)[original]
(LR-telephone,0.01395935)[original]
(LR-foreign,0.153726094)[original]
(LR-sex,-0.064938013)[ensemble]
(LR-telephone,-0.006566465)[ensemble]
(LR-foreign,0.147342564)[ensemble]
(ADA-foreign,0.198217213)[original]
(ADA-telephone,-0.01421947)[original]
(ADA-sex,-0.052757388)[original]
(ADA-foreign,0.176429092)[ensemble]
(ADA-telephone,-0.002460299)[ensemble]
(ADA-sex,-0.044927691)[ensemble]
(RF-foreign,0.151394248)[original]
(RF-sex,-0.051923281)[original]
(RF-telephone,-0.05139068)[original]
(RF-foreign,0.117447251)[ensemble]
(RF-sex,-0.046940002)[ensemble]
(RF-telephone,-0.018143304)[ensemble]
(BAGGING-foreign,0.145465808)[original]
(BAGGING-telephone,-0.037915839)[original]
(BAGGING-sex,-0.086844257)[original]
(BAGGING-foreign,0.164472569)[ensemble]
(BAGGING-telephone,0.005665585)[ensemble]
(BAGGING-sex,-0.057575597)[ensemble]

};
\end{axis}
\end{tikzpicture} 
\vspace*{-5mm}
\end{subfigure}
\begin{subfigure}[]{0.22\textwidth}
\caption{PE}
\begin{tikzpicture}
\begin{axis}[
    height=4cm,
 	width=4.5cm,
	ymax=0.45,
	ymin=-0.45,
 	xticklabel style = {font=\tiny,yshift=0ex},
 	yticklabel style = {font=\tiny},
 	x tick label style={at={(0.5,-0.25)},rotate=50,anchor=east},
 	xlabel style={yshift=-50pt},
 	ylabel style={yshift=-15pt},
    scatter/classes={original={mark=*,blue,mark size=1.5}, ensemble={mark=*,red,mark size=1.5}},
	symbolic x coords={ADA-foreign,ADA-telephone,ADA-sex,BAGGING-sex,BAGGING-foreign,BAGGING-telephone,RF-foreign,RF-telephone,RF-sex,LR-sex,LR-foreign,LR-telephone},
	xtick=data
	]
	
    
    \coordinate (A) at (axis cs:LR-sex,0);
    \coordinate (O1) at (rel axis cs:0,0);
    \coordinate (O2) at (rel axis cs:1,0);
    
    \draw [black,dashed] (A -| O1) -- (A -| O2);
    
	\addplot[scatter,only marks,scatter src=explicit symbolic] coordinates {
(LR-sex,-0.110072903)[original]
(LR-telephone,-0.205225607)[original]
(LR-foreign,0.24784025)[original]
(LR-sex,0.027612193)[ensemble]
(LR-telephone,-0.021995335)[ensemble]
(LR-foreign,0.375983468)[ensemble]
(ADA-foreign,0.409035853)[original]
(ADA-telephone,-0.111276771)[original]
(ADA-sex,-0.075603089)[original]
(ADA-foreign,0.428561839)[ensemble]
(ADA-telephone,-0.079849421)[ensemble]
(ADA-sex,0.003300779)[ensemble]
(RF-foreign,0.439250224)[original]
(RF-sex,-0.065653221)[original]
(RF-telephone,-0.095762712)[original]
(RF-foreign,0.383618352)[ensemble]
(RF-sex,-0.043314707)[ensemble]
(RF-telephone,-0.028813559)[ensemble]
(BAGGING-foreign,0.225145129)[original]
(BAGGING-telephone,-0.06568915)[original]
(BAGGING-sex,0.0315562)[original]
(BAGGING-foreign,0.391475492)[ensemble]
(BAGGING-telephone,-0.046920821)[ensemble]
(BAGGING-sex,0.015635828)[ensemble]


};

\end{axis}
\end{tikzpicture}
\vspace*{-5mm}
\end{subfigure}
\begin{subfigure}[]{0.22\textwidth}
\caption{DI}
\begin{tikzpicture}
\begin{axis}[
    height=4cm,
 	width=4.5cm,
	ymax=1.75,
	ymin=0.25,
 	xticklabel style = {font=\tiny,yshift=0ex},
 	yticklabel style = {font=\tiny},
 	x tick label style={at={(0.5,-0.25)},rotate=50,anchor=east},
 	xlabel style={yshift=-50pt},
 	ylabel style={yshift=-15pt},
    scatter/classes={original={mark=*,blue,mark size=1.5}, ensemble={mark=*,red,mark size=1.5}},
	symbolic x coords={ADA-foreign,ADA-telephone,ADA-sex,BAGGING-sex,BAGGING-foreign,BAGGING-telephone,RF-foreign,RF-telephone,RF-sex,LR-sex,LR-foreign,LR-telephone},
	xtick=data,
	legend pos=outer north east,
	legend cell align=left
	]
	
    
    \coordinate (A) at (axis cs:LR-sex,1);
    \coordinate (O1) at (rel axis cs:0,0);
    \coordinate (O2) at (rel axis cs:1,0);
    
    \draw [black,dashed] (A -| O1) -- (A -| O2);
    
	\addplot[scatter,only marks,scatter src=explicit symbolic] coordinates {
(LR-sex,0.849359271)[original]
(LR-telephone,0.864121023)[original]
(LR-foreign,1.214889379)[original]
(LR-sex,0.959621195)[ensemble]
(LR-telephone,0.97183826)[ensemble]
(LR-foreign,1.194554362)[ensemble]
(ADA-foreign,1.300982463)[original]
(ADA-telephone,0.900865707)[original]
(ADA-sex,0.891417561)[original]
(ADA-foreign,1.299226694)[ensemble]
(ADA-telephone,0.936517708)[ensemble]
(ADA-sex,0.956704502)[ensemble]
(RF-foreign,1.269206632)[original]
(RF-sex,0.892525903)[original]
(RF-telephone,0.827486983)[original]
(RF-foreign,1.169987668)[ensemble]
(RF-sex,0.91750353)[ensemble]
(RF-telephone,0.937906564)[ensemble]
(BAGGING-foreign,1.115267077)[original]
(BAGGING-telephone,0.90420676)[original]
(BAGGING-sex,0.946822337)[original]
(BAGGING-foreign,1.172875646)[ensemble]
(BAGGING-telephone,0.974407285)[ensemble]
(BAGGING-sex,0.954525823)[ensemble]


};
\end{axis}
\end{tikzpicture} 
\vspace*{-5mm}
\end{subfigure}
}
\end{center}
\caption{Fairness metrics for German Credit Score Dataset}
\label{fig:german_metrics}
\end{figure*}

\section{Conclusion and Future Work}
\label{sec:conclusion}
In this paper we revisited \F's framework that uses explanation methods in order to assess model fairness. \F uses LIME explanations, and it receives as input a triple $(M,D,F)$ of a classifier $M$, a dataset $D$ and a set of ``sensitive" features $F$, and outputs a fairer classifier $M_{final}$ in the sense that it is less dependent on sensitive features without compromising the model's accuracy. We extended the empirical study of \cite{bhargava} by including experiments of a wide family of classifiers on various and diverse datasets on which fairness issues naturally appear. These new experiments reattested what was empirically shown in \cite{bhargava}, namely, that \F improves process fairness without compromising accuracy.
\\
However, the authors of \cite{dimanov} raised several concerns in such an approach based on explanation methods that use feature importance indices to determine model fairness since they conceal other forms of unfairness. This motivated us to deepen the analysis of \F to evaluate its output models with respect to several well known fairness metrics. Our results show consistent improvements in most metrics with very few exceptions that will be investigated in more detail in the near future.
\\
We have also adapted \F to other data types and different explanatory models such as SHAP \cite{shap} and Anchors \cite{anchors}. However,  the construction of global explanations like \cite{van2019global} should be thoroughly explored. Also,
the aggregation rule to produce classifier ensembles should be improved in order take into account classifier weighting,  as well as other classifiers resulting from the removal of different subsets of sensitive features (here we only considered the removal of one or all features). Finally,
 we took a human and context-centered approach for identifying sensitive features in a given use-case. There is hope to automating this task while  taking into account domain knowledge and using statistical dataset characteristics and utility-based approaches to quantify sensitivity. This will be the topic of a follow up contribution.



\bibliographystyle{unsrt}
\bibliography{sections/references}


\appendix
\cleardoublepage
\section{Appendix}\label{sec:appendix}

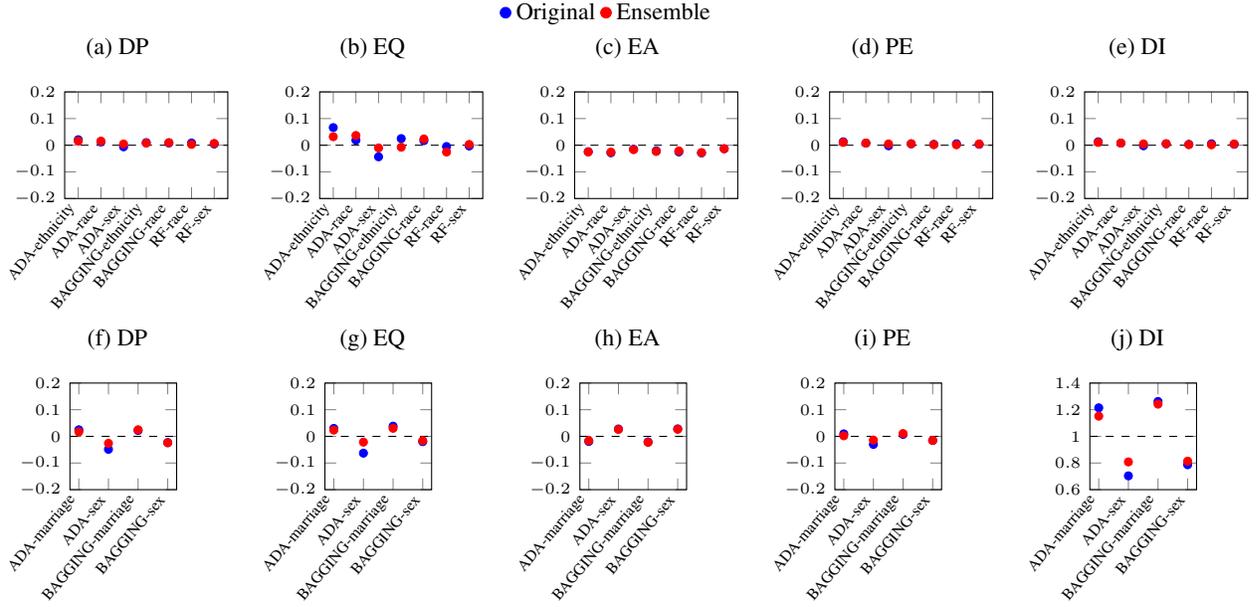
\begin{figure*}[htb]
\begin{center}
\begin{tikzpicture}[scale=1.0]
	\footnotesize
    \begin{customlegend}[legend entries={Original, Ensemble}, 
    legend columns=2, legend style={draw=none}]
    	\addlegendimage{only marks, mark=*,blue}
	    \addlegendimage{only marks, mark=*,red}
    \end{customlegend}
\end{tikzpicture}
\hbox{
\begin{subfigure}[]{0.2\textwidth}
\caption{DP}
\begin{tikzpicture}
\begin{axis}[
    height=3cm,
 	width=3.75cm,
	ymax=0.2,
	ymin=-0.2,
 	xticklabel style = {font=\tiny,yshift=0ex},
 	yticklabel style = {font=\tiny},
 	x tick label style={at={(0.5,-0.25)},rotate=50,anchor=east},
 	xlabel style={yshift=-50pt},
 	ylabel style={yshift=-15pt},
    scatter/classes={original={mark=*,blue,mark size=1.5}, ensemble={mark=*,red,mark size=1.5}},
	symbolic x coords={ADA-ethnicity,ADA-race,ADA-sex,BAGGING-ethnicity,BAGGING-race,RF-race,RF-sex},
	xtick=data
	]
	
    
    \coordinate (A) at (axis cs:ADA-ethnicity,0);
    \coordinate (O1) at (rel axis cs:0,0);
    \coordinate (O2) at (rel axis cs:1,0);
    
    \draw [black,dashed] (A -| O1) -- (A -| O2);
    
	\addplot[scatter,only marks,scatter src=explicit symbolic] coordinates {
(ADA-ethnicity,0.02062782)[original]
(ADA-race,0.01247212)[original]
(ADA-sex,-0.005957823)[original]
(ADA-ethnicity,0.015400563)[ensemble]
(ADA-race,0.015432643)[ensemble]
(ADA-sex,0.004567055)[ensemble]
(BAGGING-ethnicity,0.009707082)[original]
(BAGGING-race,0.008381722)[original]
(BAGGING-ethnicity,0.007881921)[ensemble]
(BAGGING-race,0.009845794)[ensemble]
(RF-race,0.008290422)[original]
(RF-sex,0.00464644)[original]
(RF-race,0.003084932)[ensemble]
(RF-sex,0.006745829)[ensemble]


};
\end{axis}
\end{tikzpicture}
\end{subfigure}
\begin{subfigure}[]{0.2\textwidth}
\caption{EQ}
\begin{tikzpicture}
\begin{axis}[
    height=3cm,
 	width=3.75cm,
	ymax=0.2,
	ymin=-0.2,
 	xticklabel style = {font=\tiny,yshift=0ex},
 	yticklabel style = {font=\tiny},
 	x tick label style={at={(0.5,-0.25)},rotate=50,anchor=east},
 	xlabel style={yshift=-50pt},
 	ylabel style={yshift=-15pt},
    scatter/classes={original={mark=*,blue,mark size=1.5}, ensemble={mark=*,red,mark size=1.5}},
	symbolic x coords={ADA-ethnicity,ADA-race,ADA-sex,BAGGING-ethnicity,BAGGING-race,RF-race,RF-sex},
	xtick=data
	]
	
    
    \coordinate (A) at (axis cs:ADA-ethnicity,0);
    \coordinate (O1) at (rel axis cs:0,0);
    \coordinate (O2) at (rel axis cs:1,0);
    
    \draw [black,dashed] (A -| O1) -- (A -| O2);
    
	\addplot[scatter,only marks,scatter src=explicit symbolic] coordinates {
(ADA-ethnicity,0.066157881)[original]
(ADA-race,0.018819485)[original]
(ADA-sex,-0.043420679)[original]
(ADA-ethnicity,0.031933601)[ensemble]
(ADA-race,0.036558201)[ensemble]
(ADA-sex,-0.010272578)[ensemble]
(BAGGING-ethnicity,0.024899567)[original]
(BAGGING-race,0.017253912)[original]
(BAGGING-ethnicity,-0.007464724)[ensemble]
(BAGGING-race,0.023514221)[ensemble]
(RF-race,-0.004859308)[original]
(RF-sex,-0.002842057)[original]
(RF-race,-0.025558792)[ensemble]
(RF-sex,0.002966901)[ensemble]


};
\end{axis}
\end{tikzpicture}
\end{subfigure}
\begin{subfigure}[]{0.2\textwidth}
\caption{EA}
\begin{tikzpicture}
\begin{axis}[
    height=3cm,
 	width=3.75cm,
	ymax=0.2,
	ymin=-0.2,
 	xticklabel style = {font=\tiny,yshift=0ex},
 	yticklabel style = {font=\tiny},
 	x tick label style={at={(0.5,-0.25)},rotate=50,anchor=east},
 	xlabel style={yshift=-50pt},
 	ylabel style={yshift=-15pt},
    scatter/classes={original={mark=*,blue,mark size=1.5}, ensemble={mark=*,red,mark size=1.5}},
	symbolic x coords={ADA-ethnicity,ADA-race,ADA-sex,BAGGING-ethnicity,BAGGING-race,RF-race,RF-sex},
	xtick=data
	]
	
    
    \coordinate (A) at (axis cs:ADA-ethnicity,0);
    \coordinate (O1) at (rel axis cs:0,0);
    \coordinate (O2) at (rel axis cs:1,0);
    
    \draw [black,dashed] (A -| O1) -- (A -| O2);
    
	\addplot[scatter,only marks,scatter src=explicit symbolic] coordinates {
(ADA-ethnicity,-0.024323003)[original]
(ADA-race,-0.028311873)[original]
(ADA-sex,-0.014965169)[original]
(ADA-ethnicity,-0.025307512)[ensemble]
(ADA-race,-0.0258356)[ensemble]
(ADA-sex,-0.016476898)[ensemble]
(BAGGING-ethnicity,-0.021036074)[original]
(BAGGING-race,-0.02502429)[original]
(BAGGING-ethnicity,-0.02346034)[ensemble]
(BAGGING-race,-0.021443482)[ensemble]
(RF-race,-0.029040399)[original]
(RF-sex,-0.013144597)[original]
(RF-race,-0.027711922)[ensemble]
(RF-sex,-0.013221761)[ensemble]

};
\end{axis}
\end{tikzpicture}
\end{subfigure}
\begin{subfigure}[]{0.2\textwidth}
\caption{PE}
\begin{tikzpicture}
\begin{axis}[
    height=3cm,
 	width=3.75cm,
	ymax=0.2,
	ymin=-0.2,
 	xticklabel style = {font=\tiny,yshift=0ex},
 	yticklabel style = {font=\tiny},
 	x tick label style={at={(0.5,-0.25)},rotate=50,anchor=east},
 	xlabel style={yshift=-50pt},
 	ylabel style={yshift=-15pt},
    scatter/classes={original={mark=*,blue,mark size=1.5}, ensemble={mark=*,red,mark size=1.5}},
	symbolic x coords={ADA-ethnicity,ADA-race,ADA-sex,BAGGING-ethnicity,BAGGING-race,RF-race,RF-sex},
	xtick=data
	]
	
    
    \coordinate (A) at (axis cs:ADA-ethnicity,0);
    \coordinate (O1) at (rel axis cs:0,0);
    \coordinate (O2) at (rel axis cs:1,0);
    
    \draw [black,dashed] (A -| O1) -- (A -| O2);
    
	\addplot[scatter,only marks,scatter src=explicit symbolic] coordinates {

(ADA-ethnicity,0.012919524)[original]
(ADA-race,0.007884789)[original]
(ADA-sex,-0.002084572)[original]
(ADA-ethnicity,0.010598172)[ensemble]
(ADA-race,0.008205877)[ensemble]
(ADA-sex,0.004855529)[ensemble]
(BAGGING-ethnicity,0.005151161)[original]
(BAGGING-race,0.003257135)[original]
(BAGGING-ethnicity,0.005609853)[ensemble]
(BAGGING-race,0.00224149)[ensemble]
(RF-race,0.005350827)[original]
(RF-sex,0.003495318)[original]
(RF-race,0.001644274)[ensemble]
(RF-sex,0.004758649)[ensemble]

};
\end{axis}
\end{tikzpicture}
\end{subfigure}
\begin{subfigure}[]{0.2\textwidth}
\caption{DI}
\begin{tikzpicture}
\begin{axis}[
    height=3cm,
 	width=3.75cm,
	ymax=0.2,
	ymin=-0.2,
 	xticklabel style = {font=\tiny,yshift=0ex},
 	yticklabel style = {font=\tiny},
 	x tick label style={at={(0.5,-0.25)},rotate=50,anchor=east},
 	xlabel style={yshift=-50pt},
 	ylabel style={yshift=-15pt},
    scatter/classes={original={mark=*,blue,mark size=1.5}, ensemble={mark=*,red,mark size=1.5}},
	symbolic x coords={ADA-ethnicity,ADA-race,ADA-sex,BAGGING-ethnicity,BAGGING-race,RF-race,RF-sex},
	xtick=data,
	legend pos=outer north east,
	legend cell align=left
	]
	
    
    \coordinate (A) at (axis cs:ADA-ethnicity,0);
    \coordinate (O1) at (rel axis cs:0,0);
    \coordinate (O2) at (rel axis cs:1,0);
    
    \draw [black,dashed] (A -| O1) -- (A -| O2);
    
	\addplot[scatter,only marks,scatter src=explicit symbolic] coordinates {

(ADA-ethnicity,0.012919524)[original]
(ADA-race,0.007884789)[original]
(ADA-sex,-0.002084572)[original]
(ADA-ethnicity,0.010598172)[ensemble]
(ADA-race,0.008205877)[ensemble]
(ADA-sex,0.004855529)[ensemble]
(BAGGING-ethnicity,0.005151161)[original]
(BAGGING-race,0.003257135)[original]
(BAGGING-ethnicity,0.005609853)[ensemble]
(BAGGING-race,0.00224149)[ensemble]
(RF-race,0.005350827)[original]
(RF-sex,0.003495318)[original]
(RF-race,0.001644274)[ensemble]
(RF-sex,0.004758649)[ensemble]

};
\end{axis}
\end{tikzpicture}
\end{subfigure}
}
\hbox{

\begin{subfigure}[]{0.2\textwidth}
\caption{DP}
\begin{tikzpicture}
\begin{axis}[
    height=3cm,
 	width=3cm,
	ymax=0.2,
	ymin=-0.2,
 	xticklabel style = {font=\tiny,yshift=0ex},
 	yticklabel style = {font=\tiny},
 	x tick label style={at={(0.5,-0.25)},rotate=50,anchor=east},
 	xlabel style={yshift=-50pt},
 	ylabel style={yshift=-15pt},
    scatter/classes={original={mark=*,blue,mark size=1.5}, ensemble={mark=*,red,mark size=1.5}},
	symbolic x coords={ADA-marriage,ADA-sex,BAGGING-marriage,BAGGING-sex},
	xtick=data
	]
	
    
    \coordinate (A) at (axis cs:ADA-marriage,0);
    \coordinate (O1) at (rel axis cs:0,0);
    \coordinate (O2) at (rel axis cs:1,0);
    
    \draw [black,dashed] (A -| O1) -- (A -| O2);
    
	\addplot[scatter,only marks,scatter src=explicit symbolic] coordinates {

(BAGGING-marriage,0.022030737)[original]
(BAGGING-sex,-0.023331794)[original]
(BAGGING-marriage,0.024762293)[ensemble]
(BAGGING-sex,-0.024162885)[ensemble]
(ADA-sex,-0.048702134)[original]
(ADA-marriage,0.024662045)[original]
(ADA-sex,-0.026129056)[ensemble]
(ADA-marriage,0.016943398)[ensemble]

};
\end{axis}
\end{tikzpicture}
\end{subfigure}
\begin{subfigure}[]{0.2\textwidth}
\caption{EQ}
\begin{tikzpicture}
\begin{axis}[
    height=3cm,
 	width=3cm,
	ymax=0.2,
	ymin=-0.2,
 	xticklabel style = {font=\tiny,yshift=0ex},
 	yticklabel style = {font=\tiny},
 	x tick label style={at={(0.5,-0.25)},rotate=50,anchor=east},
 	xlabel style={yshift=-50pt},
 	ylabel style={yshift=-15pt},
    scatter/classes={original={mark=*,blue,mark size=1.5}, ensemble={mark=*,red,mark size=1.5}},
	symbolic x coords={ADA-marriage,ADA-sex,BAGGING-marriage,BAGGING-sex},
	xtick=data
	]
	
    
    \coordinate (A) at (axis cs:ADA-marriage,0);
    \coordinate (O1) at (rel axis cs:0,0);
    \coordinate (O2) at (rel axis cs:1,0);
    
    \draw [black,dashed] (A -| O1) -- (A -| O2);
    
	\addplot[scatter,only marks,scatter src=explicit symbolic] coordinates {

(BAGGING-marriage,0.038188649)[original]
(BAGGING-sex,-0.019961133)[original]
(BAGGING-marriage,0.030301694)[ensemble]
(BAGGING-sex,-0.016227476)[ensemble]
(ADA-sex,-0.063055198)[original]
(ADA-marriage,0.02982448)[original]
(ADA-sex,-0.02165828)[ensemble]
(ADA-marriage,0.023404017)[ensemble]

};
\end{axis}
\end{tikzpicture}
\end{subfigure}
\begin{subfigure}[]{0.2\textwidth}
\caption{EA}
\begin{tikzpicture}
\begin{axis}[
    height=3cm,
 	width=3cm,
	ymax=0.2,
	ymin=-0.2,
 	xticklabel style = {font=\tiny,yshift=0ex},
 	yticklabel style = {font=\tiny},
 	x tick label style={at={(0.5,-0.25)},rotate=50,anchor=east},
 	xlabel style={yshift=-50pt},
 	ylabel style={yshift=-15pt},
    scatter/classes={original={mark=*,blue,mark size=1.5}, ensemble={mark=*,red,mark size=1.5}},
	symbolic x coords={ADA-marriage,ADA-sex,BAGGING-marriage,BAGGING-sex},
	xtick=data
	]
	
    
    \coordinate (A) at (axis cs:ADA-marriage,0);
    \coordinate (O1) at (rel axis cs:0,0);
    \coordinate (O2) at (rel axis cs:1,0);
    
    \draw [black,dashed] (A -| O1) -- (A -| O2);
    
	\addplot[scatter,only marks,scatter src=explicit symbolic] coordinates {

(BAGGING-marriage,-0.020791291)[original]
(BAGGING-sex,0.028014925)[original]
(BAGGING-marriage,-0.02250772)[ensemble]
(BAGGING-sex,0.026675409)[ensemble]
(ADA-sex,0.027674563)[original]
(ADA-marriage,-0.019072715)[original]
(ADA-sex,0.026138421)[ensemble]
(ADA-marriage,-0.016368538)[ensemble]

};
\end{axis}
\end{tikzpicture}
\end{subfigure}
\begin{subfigure}[]{0.2\textwidth}
\caption{PE}
\begin{tikzpicture}
\begin{axis}[
    height=3cm,
 	width=3cm,
	ymax=0.2,
	ymin=-0.2,
 	xticklabel style = {font=\tiny,yshift=0ex},
 	yticklabel style = {font=\tiny},
 	x tick label style={at={(0.5,-0.25)},rotate=50,anchor=east},
 	xlabel style={yshift=-50pt},
 	ylabel style={yshift=-15pt},
    scatter/classes={original={mark=*,blue,mark size=1.5}, ensemble={mark=*,red,mark size=1.5}},
	symbolic x coords={ADA-marriage,ADA-sex,BAGGING-marriage,BAGGING-sex},
	xtick=data
	]
	
    
    \coordinate (A) at (axis cs:ADA-marriage,0);
    \coordinate (O1) at (rel axis cs:0,0);
    \coordinate (O2) at (rel axis cs:1,0);
    
    \draw [black,dashed] (A -| O1) -- (A -| O2);
    
	\addplot[scatter,only marks,scatter src=explicit symbolic] coordinates {

(BAGGING-marriage,0.007301399)[original]
(BAGGING-sex,-0.015420975)[original]
(BAGGING-marriage,0.010466081)[ensemble]
(BAGGING-sex,-0.015351626)[ensemble]
(ADA-sex,-0.030252451)[original]
(ADA-marriage,0.009390316)[original]
(ADA-sex,-0.014090328)[ensemble]
(ADA-marriage,0.002125681)[ensemble]

};
\end{axis}
\end{tikzpicture}
\end{subfigure}
\begin{subfigure}[]{0.2\textwidth}
\caption{DI}
\begin{tikzpicture}
\begin{axis}[
    height=3cm,
 	width=3cm,
	ymax=1.4,
	ymin=0.6,
 	xticklabel style = {font=\tiny,yshift=0ex},
 	yticklabel style = {font=\tiny},
 	x tick label style={at={(0.5,-0.25)},rotate=50,anchor=east},
 	xlabel style={yshift=-50pt},
 	ylabel style={yshift=-15pt},
    scatter/classes={original={mark=*,blue,mark size=1.5}, ensemble={mark=*,red,mark size=1.5}},
	symbolic x coords={ADA-marriage,ADA-sex,BAGGING-marriage,BAGGING-sex},
	xtick=data,
	legend pos=outer north east,
	legend cell align=left
	]
	
    
    \coordinate (A) at (axis cs:ADA-marriage,1);
    \coordinate (O1) at (rel axis cs:0,0);
    \coordinate (O2) at (rel axis cs:1,0);
    
    \draw [black,dashed] (A -| O1) -- (A -| O2);
    
	\addplot[scatter,only marks,scatter src=explicit symbolic] coordinates {

(BAGGING-marriage,1.261757711)[original]
(BAGGING-sex,0.786815283)[original]
(BAGGING-marriage,1.241715624)[ensemble]
(BAGGING-sex,0.813990525)[ensemble]
(ADA-sex,0.703653371)[original]
(ADA-marriage,1.214231897)[original]
(ADA-sex,0.808161061)[ensemble]
(ADA-marriage,1.151666367)[ensemble]

};
\end{axis}
\end{tikzpicture}
\end{subfigure}
}
\end{center}

\caption{Fairness metrics for the HMDA dataset (first row) and the Default dataset (second row). For both datasets, lesser original models were deemed unfair, namely, ADA, Bagging and RF on HMDA, and ADA and Bagging on Default. Even though these models were deemed unfair by \F, most of the fairness metrics actually indicate a rather fair behaviour by the original and \F's ensemble models. 
}
\label{fig:}
\end{figure*}

\end{document}